\documentclass{amia}
\usepackage{tabularx}
\usepackage{multirow}
\usepackage{booktabs}
\usepackage{graphicx}

\usepackage[colorlinks=true, urlcolor=blue, linkcolor=red]{hyperref}
\begin{document}

\title{Improving Large Language Models for Clinical Named Entity Recognition via Prompt Engineering }

\author{Yan Hu, MS$^1$, Qingyu Chen, PhD$^{2^3}$, Jingcheng Du, PhD$^1$, Xueqing Peng, PhD$^2$, Vipina Kuttichi Keloth, PhD$^2$, Xu Zuo, MS$^1$, Yujia Zhou, MS$^1$, Zehan Li, MS$^1$, Xiaoqian Jiang, PhD$^1$, Zhiyong Lu, PhD$^3$, Kirk Roberts, PhD$^1$, Hua Xu$^2$ }

\institutes{
    $^1$ School of Biomedical Informatics, University of Texas Health Science at Houston, Houston, USA \\
    $^2$ Section of Biomedical Informatics and Data Science, School of Medicine, Yale University, New Haven, USA \\
    $^3$ National Center for Biotechnology Information, National Library of Medicine, National Institutes of Health, Maryland, USA
}

\maketitle

\section*{Abstract}

\textbf{Objective:} This study quantifies the capabilities of GPT-3.5 and GPT-4 for clinical named entity recognition (NER) tasks and proposes task-specific prompts to improve their performance. \\
\textbf{Materials and Methods:} We evaluated these models on two clinical NER tasks:  (1) to extract medical problems, treatments, and tests from clinical notes in the MTSamples corpus, following the 2010 i2b2 concept extraction shared task, and (2) identifying nervous system disorder-related adverse events from safety reports in the vaccine adverse event reporting system (VAERS). To improve the GPT models' performance, we developed a clinical task-specific prompt framework that includes (1) baseline prompts with task description and format specification, (2) annotation guideline-based prompts, (3) error analysis-based instructions, and (4) annotated samples for few-shot learning. We assessed each prompt's effectiveness and compared the models to BioClinicalBERT. \\
\textbf{Results:} Using baseline prompts, GPT-3.5 and GPT-4 achieved relaxed F1 scores of 0.634, 0.804 for MTSamples, and 0.301, 0.593 for VAERS. Additional prompt components consistently improved model performance. When all four components were used,  GPT-3.5 and GPT-4 achieved relaxed F1 socres of 0.794, 0.861 for MTSamples and 0.676, 0.736 for VAERS, demonstrating the effectiveness of our prompt framework. Although these results trail BioClinicalBERT (F1 of 0.901 for the MTSamples dataset and 0.802 for the VAERS), it is very promising considering few training samples are needed. \\
\textbf{Conclusion:} While direct application of GPT models to clinical NER tasks falls short of optimal performance, our task-specific prompt framework, incorporating medical knowledge and training samples, significantly enhances GPT models' feasibility for potential clinical applications.

\section*{1 Introduction}

Electronic health records (EHRs) contain a vast quantity of unstructured data, including clinical notes, which can offer valuable insights for patient care and clinical research \cite{jensen2012mining}. However, manually extracting pertinent information from clinical notes presents a challenge, as it is labor-intensive and time-consuming. To address these challenges, researchers have developed various natural language processing (NLP) techniques for automating the clinical information extraction process. Clinical named entity recognition (NER) is a critical clinical NLP task focusing on recognizing boundaries of clinical entities (i.e., words/phrases) and determining their semantic categories, such as medical problems, treatment, and tests \cite{nadkarni2011natural}. With the help of advancements in clinical NER, the time and effort required for manual chart review and coding by health professionals can be significantly reduced, thus improving patient care efficiency and accelerating clinical research \cite{neveol2018clinical}.

Early clinical NER systems are often dependent on predefined lexical resources and syntactic/semantic rules derived from extensive manual analysis of text \cite{wang2018clinical}. Over the past decade, machine learning-based approaches have gained popularity in clinical NER research \cite{huang2015bidirectional}. Current popular clinical information extraction systems, such as cTAKES and CLAMP, are hybrid systems that integrate rule-based and machine learning-based techniques \cite{savova2010mayo}. Nevertheless, a bottleneck in building machine learning-based clinical NER models is to develop large, annotated corpora, which often require domain experts and take a long time to build. More recently, transformer-based large language models have emerged as the leading method for developing clinical NLP applications. Bidirectional Encoder Representations from Transformers (BERT) is a widely used pre-trained language model that learns contextual representations of free text \cite{devlin2018bert}. Utilizing BERT as the foundation, domain-specific language models like BioBERT, PubMedBERT (trained on biomedical literature), and ClinicalBERT (trained on the MIMIC-III dataset) have been further developed \cite{lee2020biobert,gu2021domain,huang2019clinicalbert}. These models have been applied to clinical NER tasks via transfer learning (i.e., fine-tuning the models on clinical NER corpora), and have shown improved performance with fewer annotated samples \cite{lee2020biobert,gu2021domain,huang2019clinicalbert}.

Generative Pre-trained Transformers (GPT) represent another type of large language model capable of generating human-like responses based on textual input. In November 2022, OpenAI unveiled GPT-3.5 \cite{openai}, a groundbreaking chatbot driven by the GPT-3.5 language model that quickly garnered interest from researchers and technology enthusiasts. As an extension of GPT-3, GPT-3.5 serves as a conversational agent adept at following complex instructions and generating high-quality responses across various scenarios. Besides its conversational skills, GPT-3.5 has exhibited remarkable performance in many other NLP tasks, such as machine translation and question-answering \cite{bang2023multitask}, even in the zero-shot or few-shot learning scenarios \cite{brown2020language}, where the model can be applied to new tasks without any fine-tuning or with fine-tuning using a very small amount of data. On March 18th, 2023, OpenAI released GPT-4, one of the most advanced NLP models at the time, which has demonstrated even greater capabilities and performance improvements over GPT-3.5 \cite{achiam2023gpt}.

As interest in GPT models continues to surge, numerous studies are currently exploring the wide range of possibilities offered by these large language models. One prominent example of GPT models for medicine is that GPT-3.5 passed the US medical license exam with about 60\% accuracy, which has further sparked the potential use of GPT-3.5 and GPT-4 in the medical domain \cite{gilson2023does}. More applications of GPT-3.5 and GPT-4 in healthcare have also been discussed \cite{kung2023performance,rao2023evaluating,antaki2023evaluating,jeblick2022chatgpt,peter_goldbert_kohane_2023,chen2023large,tian2024opportunities,jin2023genegpt}. With those motivations, this study aims to investigate the potential of GPT models for clinical NER tasks.

Meanwhile, prompt engineering has emerged as a crucial aspect of utilizing GPT models effectively for various NLP tasks. Prompt engineering involves designing input prompts that guide the model to generate desired outputs, thereby improving its performance on specific tasks \cite{wang2023prompt}. Several studies have explored prompt engineering for GPT models in open-domain settings, demonstrating its effectiveness in enhancing the model's performance across a range of tasks \cite{yu2023exploring,ma2023prompt}. In the biomedical domain, some work has been done on prompt engineering for GPT models, focusing on tasks such as biomedical question-answering, text classification and NER \cite{hsueh2023ncu,ateia2023chatgpt,chen2023evaluation}. However, to the best of our knowledge, no work has been conducted on prompt engineering for GPT models specifically targeting NER tasks in clinical texts. This highlights the need for further investigation into the potential of GPT models and prompt engineering techniques for clinical NER applications.

The contributions of this study are threefold. First, we proposed a prompt framework for clinical NER by incorporating entity definitions, annotation guidelines, and annotated samples, and demonstrated its effectiveness on two NER tasks (e.g., improving the performance of the GPT models by up to ~20\% and making is more competitive to fine-tuned models such as BioClinicalBERT).  Second, we discussed how the recent LLMs such as GPT models will change the development of NER systems in the medical domain. This is important because LLMs shows a great potential for developing generalizable clinical NER systems without substantial annotation efforts. Finally, this study also established a novel benchmark to evaluate the performance of the LLMs, GPT-3.5 and GPT-4, for the task of clinical NER. We leveraged two distinct clinical NER tasks as benchmarks, namely the 2010 i2b2 concept extraction task \cite{uzuner20112010} and the nervous system disorder-related event extraction task \cite{du2021extracting}. All code and datasets are made publicly available to the community. 

\section*{2 Methods}
\subsection*{2.1 Task Overview}
This study aims to assess the zero-shot capability of GPT-3.5 based ChatGPT (in the following discussion, we will use ChatGPT to refer to GPT-3.5 based ChatGPT) in the clinical NER task, as defined in the 2010 i2b2 challenge \cite{uzuner20112010}. We compared the performance of ChatGPT and GPT-3 in a similar zero-shot setting and included a baseline model, BioClinicalBERT, which was trained on the 2010 i2b2 dataset detailed below. The primary workflow of our investigation is depicted in Figure~\ref{fig:workflow}. Two different prompts were crafted to identify three types of clinical entities: Medical Problem, Treatment, and Test from clinical text using both ChatGPT and GPT-3. Additionally, we trained a supervised  BioClinicalBERT model using an annotated corpus from the 2010 i2b2 challenge, as a baseline. All three models were then evaluated using an annotated corpus consisting of HPI sections from 100 discharge summaries in the MTSamples collection (see next section).

\begin{figure}[H]
  \centering
    \includegraphics[width=1.0\textwidth]{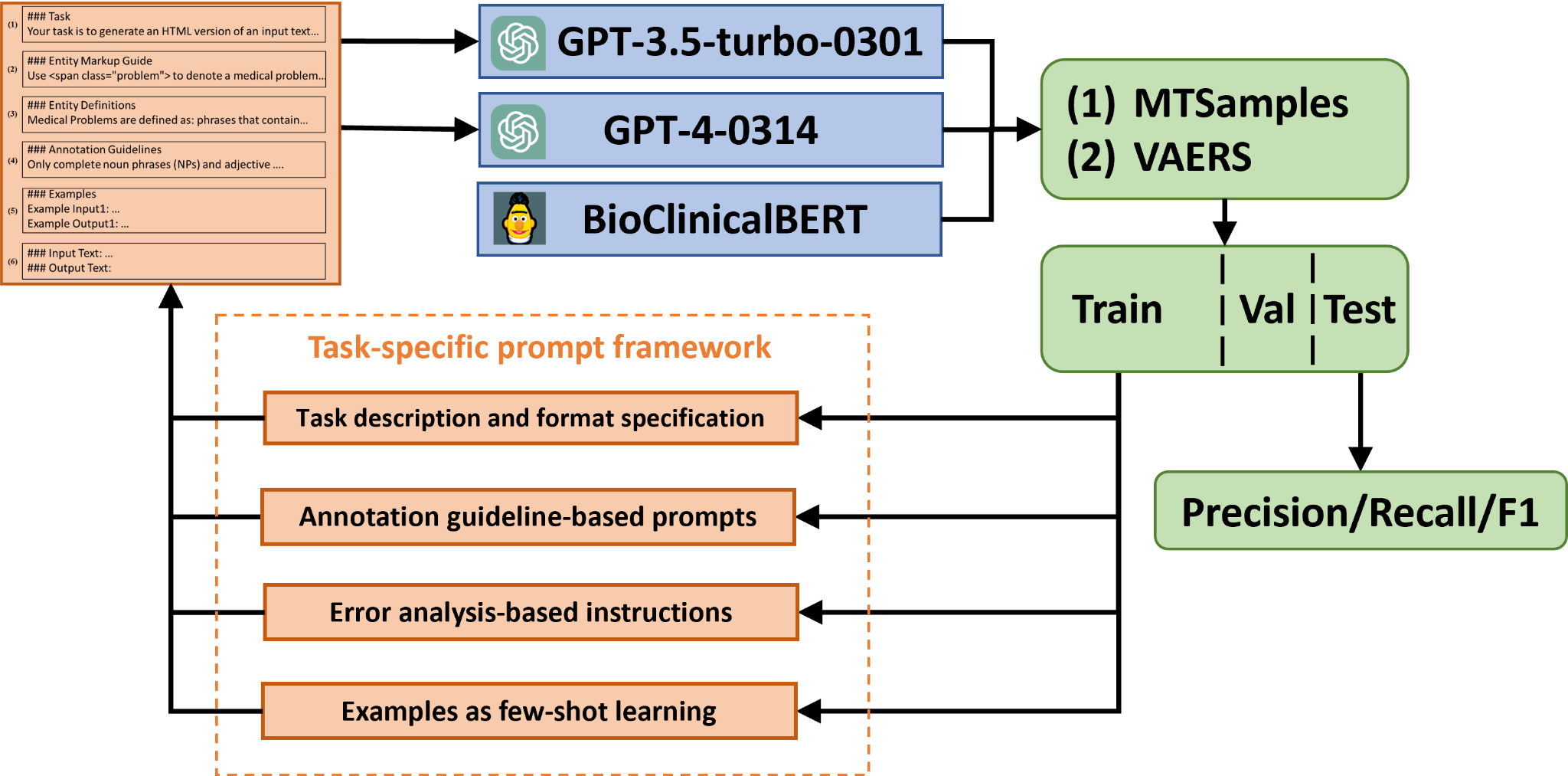}
 \caption{An overview of the study workflow.}
 \label{workflow.png}
\end{figure}

\subsection*{2.2 Dataset}
Two clinical NER datasets were used in our study, including (1) MTSamples, a set of 163 fully synthetic discharge summaries from MTSamples, which was annotated according to the annotation guidelines from the 2010 i2b2 challenge, which aims at extracting Medical Problem, Treatment, and Test \cite{uzuner20112010}; (2) the VAERS corpus, a set of 91 publicly available safety reports in VAERS, aiming at extracting nervous system disorder-related events \cite{du2021extracting}. 
The MTSamples dataset is fully synthetic, meaning that it has been artificially generated and contains no real patient information. The VAERS dataset, on the other hand, is derived from publicly available post-market safety reports that are anonymized and do not contain personally identifiable information. So, no sensitive data was sent to OpenAI API, making this study free of privacy concerns. Upon consultation, it was determined that our study did not require an IRB approval.
The two datasets were split into training, validation, and test subsets. The training and validation subsets served the purpose of fine-tuning the BioClinicalBERT model. Annotated samples in prompts were randomly sampled from the training sets. The training sets were also used for error analysis to optimize our prompt strategies. The test subsets, however, were reserved exclusively for evaluating the final performance and for comparative analysis. A descriptive statistic of entities in these datasets is presented in Table~\ref{tab:stats}.

\begin{table}[h]
\centering
\caption{Dataset statistics utilized in this study.}
\begin{tabular}{|l|l|l|l|l|l|}
\hline
Datasets                   & Entities                & Train & Valid & Test & Total \\ \hline
\multirow{3}{*}{MTSamples} & Medical   Problem       & 538   & 203   & 199  & 940   \\ \cline{2-6} 
                           & Treatment               & 149   & 43    & 35   & 227   \\ \cline{2-6} 
                           & Test                    & 120   & 39    & 50   & 209   \\ \hline
\multirow{4}{*}{VAERS}     & Investigation           & 148   & 29    & 59   & 236   \\ \cline{2-6} 
                           & Nervous   adverse event & 406   & 83    & 162  & 651   \\ \cline{2-6} 
                           & Other   adverse event   & 301   & 62    & 167  & 530   \\ \cline{2-6} 
                           & Procedure               & 338   & 57    & 126  & 521   \\ \hline
\end{tabular}
\label{tab:stats}
\end{table}

\subsection*{2.3 Models}
We fine-tuned NER models using BioClinicalBERT \cite{alsentzer2019publicly}, to serve as baselines of traditional supervised learning approaches. We present results for supervised learning on both the MTSamples test set and the VAERS test set. The model weights were initialized using the transformers package, available at huggingface\footnote{\url{https://huggingface.co/emilyalsentzer/Bio_ClinicalBERT}} \cite{wolf2020transformers}. The hyperparameters employed during model training included a learning rate of 5e-5, a training batch size of 4, 20 epochs, and a weight decay of 0.01 using the AdamW optimizer \cite{loshchilov2017decoupled}.  In addition to fine-tuning NER models using BioClinicalBERT, we also employed a traditional machine learning approach for comparison. We utilized a Conditional Random Field (CRF) model with word features, including Bag-of-word, capitalization of letters in words, and prefixes and suffixes of words \cite{jiang2011study}.\\
Regarding the GPT models, we used the specific versions GPT-3.5-turbo-0301 and GPT-4-0314 for reproducibility. Temperature in a generative language model refers to a parameter that controls the randomness in the model's predictions, typically ranging from 0 (completely deterministic) to 1 or higher (increasingly random and diverse outputs). The temperature parameter for GPT models was set to 0 to minimize randomness in response generation. A lower temperature value restricts the model's tendency to take creative leaps, thereby ensuring more predictable and consistent outputs. This is crucial in clinical NER tasks where accuracy and reliability of information extraction are paramount. In our setup, the GPT models were interacted with in a 'user' role. This role simulates a real-world user interaction with the model, where the 'user' inputs prompts and the model generates responses accordingly. This approach reflects a typical use-case scenario for these models in practical applications. All input and output datasets along with prompt variants are included with Jypter notebooks that can interface with the OpenAI API in our GitHub repository.  At the time of this study, costs of GPT-3.5 per 1k tokens were approximately \$0.03 for input and \$0.06 for output. Costs of GPT-4 per 1k tokens were approximately \$ 0.001 for input and \$ 0.002 for output.  Because of privacy issues, notes containing Personal Identifiable Information (PII) could not be used in this experiment and should not be used with the GPT API.

\subsection*{2.4 Prompt engineering}
For GPT models, we proposed a task-specific prompt including the following components:\\
\begin{quotation}
(1)	Baseline prompt with task description and format specification: This component provides the LLMs with basic information about the tasks we are instructing them to perform and in what format the LLMs should output results. We instructed the models to highlight the named entities within an HTML file using \textless span\textgreater  tags with a class attribute indicating the entity types. This allows the output from GPT models to be easily converted into a traditional Inside-Outside-Beginning (IOB) format, which allows for a direct comparison of NER performance with findings from existing studies.\\
\end{quotation}
\begin{quotation}
(2)	Annotation guideline-based prompts: This component contains entity definitions and linguistic rules derived from annotation guidelines. Entity definitions offer comprehensive, unambiguous descriptions of an entity within the context of a given task. They play an instrumental role in steering the LLM toward the precise identification of entities within text documents. We noticed that the model's predictions often differed substantially from the gold standard in terms of grammatical structure. For example, discrepancies may arise concerning what types of phrases to be included (e.g., noun phrases or adjective phrases). To enhance the model's performance, we referred to and incorporated rules in the annotation guidelines to address these issues.\\
\end{quotation}
\begin{quotation}
(3)	Error analysis-based instructions: In addition to the original annotation guidelines, we also incorporated additional guidelines following error analysis of GPT outputs using the training data. For example, we noticed that GPT models often tend to annotate consultation procedures as test entities. To prevent this, we incorporated a specific rule stating, "Consultation procedures should not be annotated as tests.".\\
\end{quotation}
\begin{quotation}
(4)	Annotated samples: To further assist the LLMs in understanding the task and generating accurate results, we provided a set of annotated samples to improve its performance in a few-shot learning setting. We randomly selected either 1 or 5 annotated examples (1 or 5-shot learning) from the training set and formatted them according to the task description and entity markup guide. 
For instance, given a sentence ‘He had been diagnosed with osteoarthritis of the knees and had undergone arthroscopy years prior to admission .’ with ‘osteoarthritis of the knees’ and ‘arthroscopy’ annotated as medical problem and test entities , we incorporated this sentence into the prompt using the following format:\\
\#\#\# Examples\\
Example Input: He had been diagnosed with osteoarthritis of the knees and had undergone arthroscopy years prior to admission .  \\
Example Output: He had been diagnosed with \textless span class="problem"\textgreater osteoarthritis of the knees\textless /span\textgreater  and had undergone \textless span class="test"\textgreater arthroscopy\textless /span\textgreater  years prior to admission .  \\
\end{quotation}
We compared the effectiveness of different prompt components by incrementally incorporating annotation guideline-based prompts, error analysis-based instructions and annotated samples as shown in Table 2 (see the complete prompts for two datasets in supplementary materials S1.1).\\

\begin{table}[h]
\begin{tabularx}{\textwidth}{p{0.15\textwidth}|X}
\toprule
Prompt Types	            & Examples \\
\hline
\multirow{3}{8em}{(1) Baseline prompts} & \#\#\# Task\\
 & Your task is to generate an HTML version of an input text, marking up specific entities related to healthcare. The entities to be identified are: 'medical problems', 'treatments', and 'tests'. Use HTML \textless span\textgreater tags to highlight these entities. Each \textless span\textgreater should have a class attribute indicating the type of the entity. \\
 & \\ 
 & \#\#\# Entity Markup Guide \\
 & Use \textless span class="problem"\textgreater to denote a medical problem… \\
\hline
 \multirow{4}{8em}{(2) Annotation guideline-based prompts} & \#\#\# Task \\
 & Medical Problems are defined as: phrases that contain observations made by patients or clinicians about the patient’s body or mind that are thought to be abnormal or caused by a disease… \\
 & \\ 
 & \#\#\# Annotation Guidelines: \\
 & Only complete noun phrases (NPs) and adjective phrases (APs) should be marked. Terms that fit concept semantic rules, but that are only used as modifiers in a noun phrase should not be marked… \\
\hline
\multirow{3}{8em}{(3) Error analysis-based instructions} & \#\#\# Error-analysis-based Guidelines: \\
& Consultation procedures should not be annotated as tests… \\
\\
\hline
\multirow{3}{8em}{(4) Annotated samples via few-shot learning} & \#\#\# Examples \\
& Example Input1: He had been diagnosed with osteoarthritis of the knees and had undergone arthroscopy years prior to admission .  \\
& Example Output1: He had been diagnosed with \textless span class="problem"\textgreater osteoarthritis of the knees\textless /span\textgreater  and had undergone \textless span class="test"\textgreater arthroscopy\textless /span\textgreater  years prior to admission…\\
\bottomrule
\end{tabularx}
\caption{An Illustration of the prompt framework for clinical NER.}
\label{tab:Prompts}
\end{table}

\subsection*{2.5 Evaluation}
The performance of the models was evaluated using Precision (P), Recall (R), and F1 scores, following the same evaluation script in the 2010 i2b2 challenge \cite{uzuner20112010}. These scores were computed based on both exact-match and relaxed-match criteria. In the context of an exact match, an extracted entity should have identical token boundary and entity type as that in the gold standard. For relaxed-match, an extracted entity that exhibits overlap in text and shares the same entity type with the gold standard is acceptable.

\section*{3 Results}
\subsection*{3.1 Zero-shot performance with different prompts}
The performance evaluation of GPT-3.5 and GPT-4 in zero-shot settings using different prompts are detailed in Table 3 and Figure 2. Following the integration of annotation guideline-based prompts and error analysis-based instructions, we noticed an improvement in the performance metrics of both GPT models, across each dataset and under each evaluation criteria. Interestingly, we found these two components to have a more pronounced effect on the performance of GPT-3.5 than on GPT-4. More specifically, GPT-3.5 demonstrated an average increase of 0.09 in overall F1 scores, ranging from 0.04 to 0.14. Conversely, GPT-4 displayed a more restrained average improvement of 0.06, with a range of 0.01 to 0.10. Looking at the dataset-specific effects, these two components had a more substantial impact on the VARES dataset compared to the MTSamples dataset. For VARES, we saw an average increase of approximately 0.11, with a range from 0.09 to 0.14. In contrast, for MTSamples, we saw a more modest approximate average increase of 0.04, with the range extending from 0.01 to 0.08.

\begin{table}[H]
\begin{tabular}{m{3em}m{5em}lllllllllllll}
\toprule
& &\multicolumn{6}{c}{MTSamples} &\multicolumn{6}{c}{VAERS} \\
\cmidrule(lr){3-8}\cmidrule(lr){9-14}
& &\multicolumn{3}{c}{Exact-Match} &\multicolumn{3}{c}{Relaxed-match} & \multicolumn{3}{c}{Exact-Match} &\multicolumn{3}{c}{Relaxed-match} \\
\cmidrule(lr){3-5}\cmidrule(lr){6-8}\cmidrule(lr){9-11}\cmidrule(lr){12-14}
Models &Prompt Strategies &P &R &F1 &P &R &F1 &P &R &F1 &P &R &F1 \\
\midrule
\multirow{3}{8em}{GPT-3.5} & (1) &0.492&0.327&0.393&0.794&0.528&0.634&0.510&0.146&0.227&0.626&0.187&0.288 \\
 &(1)+(2) &0.453&0.405&0.428&0.736&0.680&0.707&0.575&0.200&0.297&0.687&0.243&0.359 \\
 &(1)+(2)+(3) &0.462&0.412&0.436&0.755&0.687&0.719&0.569&0.233&0.331&0.730&0.305&0.431 \\
\midrule
\multirow{3}{8em}{GPT-4} &(1)&0.486&0.546&0.514&0.762&0.852&0.804&0.420&0.397&0.408&0.599&0.568&0.583 \\
 &(1)+(2) &0.478&0.577&0.523&0.752&0.919&0.827&0.559&0.444&0.495&0.743&0.593&0.660 \\
 &(1)+(2)+(3) &0.488&0.570&0.526&0.777&0.908&0.838&0.536&0.469&0.500&0.727&0.650&0.686 \\
\bottomrule
\end{tabular}
\caption{Performance of BioClinicalBERT and zero-shot performance of ChatGPT and GPT-3 on MTSamples dataset.}
\label{tab:performance1}
\end{table}

\begin{figure}[H]
  \centering
    \includegraphics[width=0.8\textwidth]{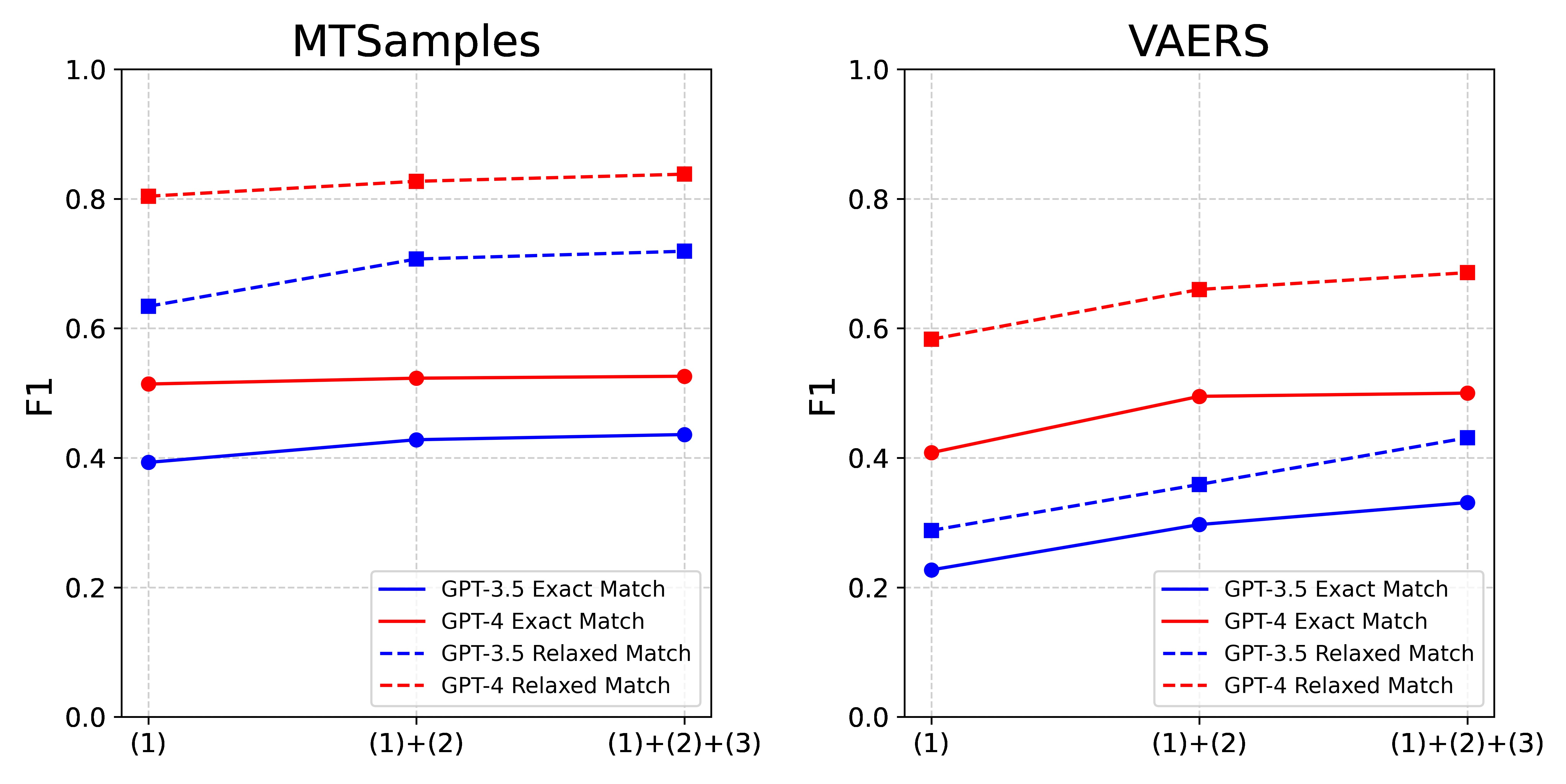}
 \caption{Performance comparison using different prompt strategies.}
 \label{fig:figure2_comparing_components}
\end{figure}

\subsection*{3.2 Effect of N-Shot Examples on Model Performance}
Table 4  and Figure 3 illustrate the performance comparison among different numbers of N-shot examples with all prompt components included. Generally, the inclusion of more examples leads to better model performance. A combination of 5-shot and all prompts produced the best results by GPT-4, achieving F1 0.593 and 0.861 for MTSamples and 0.542, 0.736 for VAERS under exact- and relaxed-match respectively. 

\begin{table}[H]
\begin{tabular}{m{3em}m{4em}lllllllllllll}
\toprule
& &\multicolumn{6}{c}{MTSamples} &\multicolumn{6}{c}{VAERS} \\
\cmidrule(lr){3-8}\cmidrule(lr){9-14}
& &\multicolumn{3}{c}{Exact-Match} &\multicolumn{3}{c}{Relaxed-match} & \multicolumn{3}{c}{Exact-Match} &\multicolumn{3}{c}{Relaxed-match} \\
\cmidrule(lr){3-5}\cmidrule(lr){6-8}\cmidrule(lr){9-11}\cmidrule(lr){12-14}
Models &Prompt Strategies &P &R &F1 &P &R &F1 &P &R &F1 &P &R &F1 \\
\midrule
\multirow{3}{8em}{GPT-3.5} & 0-shot &0.462&0.412&0.436&0.755&0.687&0.719&0.569&0.233&0.331&0.73&0.305&0.431 \\
 & 1-shot &0.475&0.461&0.468&0.779&0.778&0.779&0.561&0.311&0.401&0.733&0.416&0.531\\
 & 5-shot &0.515&0.472&0.493&0.827&0.764&0.794&0.526&0.432&0.474&0.735&0.626&0.676\\
\midrule
\multirow{3}{8em}{GPT-3.5} & 0-shot &0.488&0.570&0.526&0.777&0.908&0.838&0.536&0.469&0.500&0.727&0.650&0.686 \\
 & 1-shot &0.506&0.560&0.532&0.809&0.894&0.849&0.547&0.500&0.522&0.721&0.661&0.690\\
 & 5-shot &0.555&0.637&0.593&0.804&0.926&0.861&0.513&0.574&0.542&0.701&0.774&0.736\\
\bottomrule
\end{tabular}
\caption{0-, 1- and 5-shot performance of GPT-3.5-turbo-0301 and GPT-4-0314 using all prompt components.}
\label{tab:performance2}
\end{table}

\begin{figure}[H]
  \centering
    \includegraphics[width=0.8\textwidth]{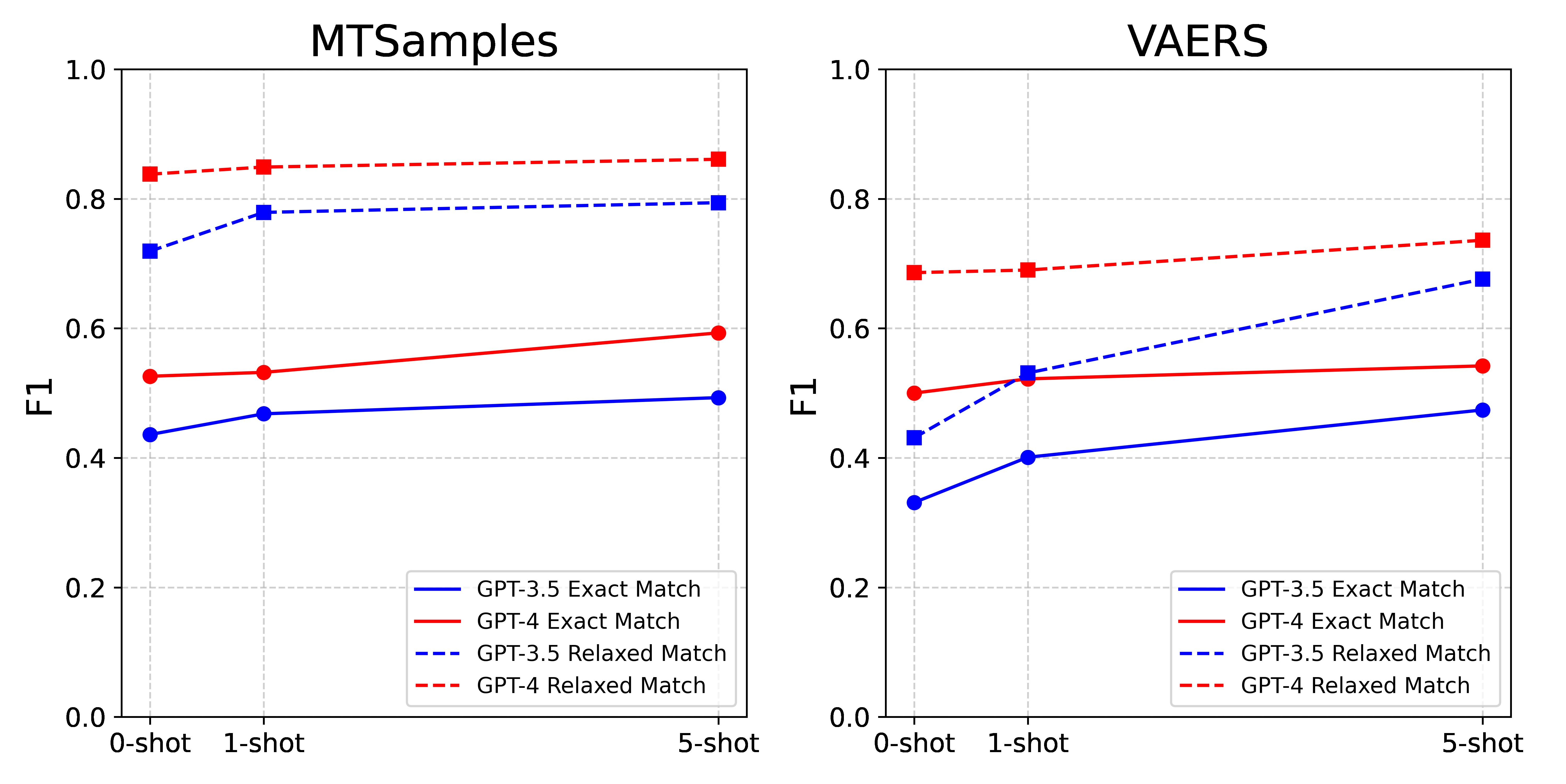}
 \caption{Performance comparison based on different numbers of N-shot examples in each prompt design.}
 \label{fig:figure3_comparing_n}
\end{figure}

\subsection*{3.3 Performance Comparison to Supervised Learning}
Table 5 displays the performance of BioClinicalBERT, CRF, GPT-3.5, and GPT-4 models for comparison. Among the three models, BioClinicalBERT still demonstrated the highest performance. For MTSamples, it achieved overall F1 scores of 0.785 and 0.901 under exact-match and relaxed-match respectively. Its performance on the VAERS dataset also remained dominant, with overall F1 scores of 0.668 and 0.802 under exact-match and relaxed-match respectively. The CRF model achieved an F1 score of 0.584 and 0.525 in MTSamples and VAERS by exact match and surpassing GPT-3.5. In the relaxed-match criteria, the CRF model performed worse than GPT-4 and GPT-3.5 in the MTSamples and had comparable performance to GPT-3.5 in the VAERS dataset. Comparatively, GPT-3.5 lagged on two datasets with the lowest performance, yet still demonstrated a decent performance with scores of 0.794 and 0.676, as evaluated by relaxed-match criteria on the two datasets respectively. GPT-4 showcased highly competitive performance using the relaxed match criteria, accomplishing F1 scores of 0.861 and 0.736 on the MTSamples and VAERS datasets respectively. It is notable, however, that the performances of GPT-3.5 and GPT-4 as evaluated by the exact-match method were not as impressive as those by the relaxed-match. In addition to the test set results, we have provided the model's performance on the validation sets in supplementary materials S1.2 to ensure the BioClinicalBERT is not overfitting.

\begin{table}[H]
\begin{tabular}{m{7em}lllllllllllll}
\toprule
& \multicolumn{6}{c}{MTSamples} &\multicolumn{6}{c}{VAERS} \\
\cmidrule(lr){2-7}\cmidrule(lr){8-13}
& \multicolumn{3}{c}{Exact-Match} &\multicolumn{3}{c}{Relaxed-match} & \multicolumn{3}{c}{Exact-Match} &\multicolumn{3}{c}{Relaxed-match} \\
\cmidrule(lr){2-4}\cmidrule(lr){5-7}\cmidrule(lr){8-10}\cmidrule(lr){11-13}
Models &P &R &F1 &P &R &F1 &P &R &F1 &P &R &F1 \\
\midrule
GPT-3.5 &0.515&0.472&0.493&0.827&0.764&0.794&0.526&0.432&0.474&0.735&0.626&0.676 \\
GPT-4 & 0.555&0.637&0.593&0.804&0.926&0.861&0.513&0.574&0.542&0.701&0.774&0.736\\
CRF & 0.511&0.681&0.584&0.662&0.887&0.758&0.473&0.591&0.525&0.609&0.764&0.678\\
BioClinicalBERT &0.785&0.785&0.785&0.915&0.887&0.901&0.698&0.640&0.668&0.846&0.761&0.802\\
\bottomrule
\end{tabular}
\caption{Performance of BioClinicalBERT, CRF, GPT-3.5, and GPT-4 on MTSamples and VAERS datasets. The performance is shown in the order of Precision/Recall/F1.}
\label{tab:performance2}
\end{table}

\begin{figure}[H]
  \centering
    \includegraphics[width=0.8\textwidth]{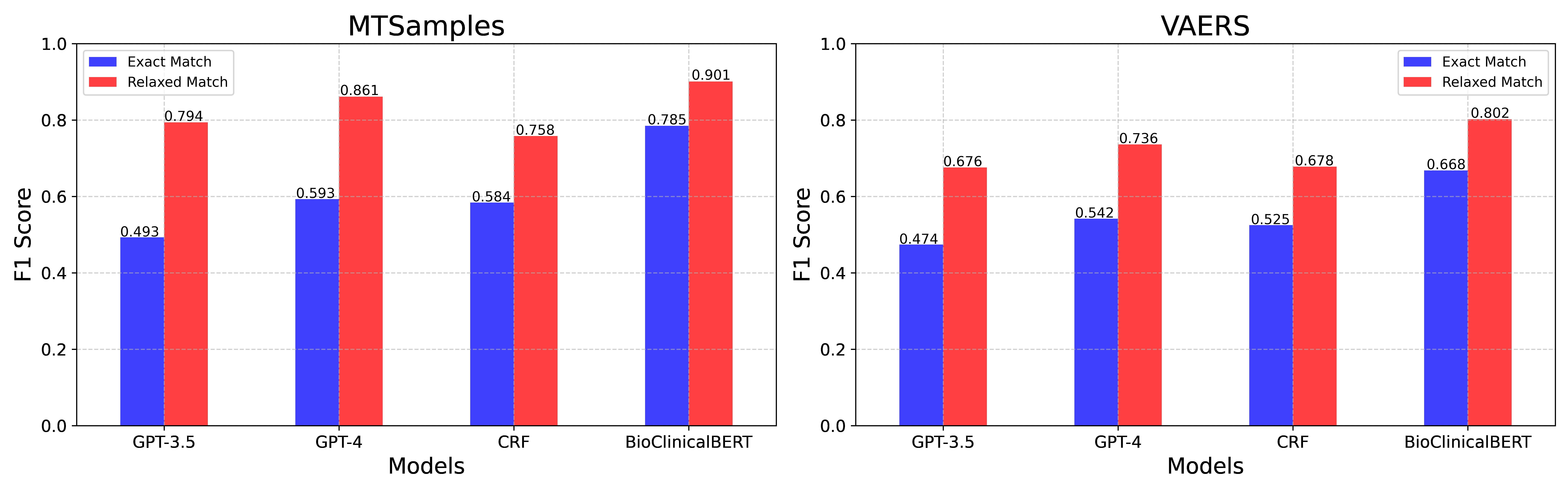}
 \caption{Performance comparison of GPT-3.5 and GPT-4 to BioClinicalBERT.}
 \label{fig:figure4_comparing_models}
\end{figure}

\subsection*{3.4 Error analysis}
A random sample of 20 sentences was selected from the outputs generated by each GPT model across the two datasets, post-processing. This selection included sentences with both false positives and false negatives. The error analysis was conducted based on exact match. The error statistics derived from this analysis are presented in Figure 5. When assessed on a dataset basis, GPT-3.5 and GPT-4 exhibited similar error patterns for the MTSamples dataset. Both models encountered challenges when it came to identifying correct entity boundaries. This typically involved making decisions on whether to include article words (such as ‘the’ in the phrase ‘the study drug’) or modifiers (such as ‘another large’ in the phrase ‘another large stroke’) that precede a noun phrase. In assessing model performance, we considered the exact-match criteria, which may present a different challenge for GPT models compared to BioClinicalBERT. While BioClinicalBERT is fine-tuned specifically on annotated entities with clear boundaries, the GPT models, being large language models, are trained on a broader and more diverse corpus. This distinction could impact their ability to adhere strictly to the exact boundaries of entities as defined in the training data, especially in the context of clinical NER where the linguistic structure and terminology are highly specialized.
As for the VAERS dataset, several factors may contribute to its increased complexity. Firstly, inner-annotator agreement was lower compared to the MTSamples dataset (i.e., average F1 0.7707 \cite{du2021extracting} vs 0.8620), indicating less consistency in annotations. Additionally, the VAERS dataset contains more semantically specific annotation categories, such as distinguishing between different types of adverse events. This specificity demands a higher level of contextual understanding from the models. On the other hand, GPT-4's major difficulties lie in determining the correct entity boundaries and accurately classifying the entity types. This discrepancy can be attributed to the unique characteristics of each dataset. The VAERS dataset contains more complex entities (i.e., Nervous adverse events vs Other adverse events) compared to the MTSamples dataset, leading to a higher error rate in entity type classification for the models. Another possible reason could be the inconsistency \cite{du2021extracting} in annotation, which needs further investigation.

\begin{figure}[H]
  \centering
    \includegraphics[width=0.8\textwidth]{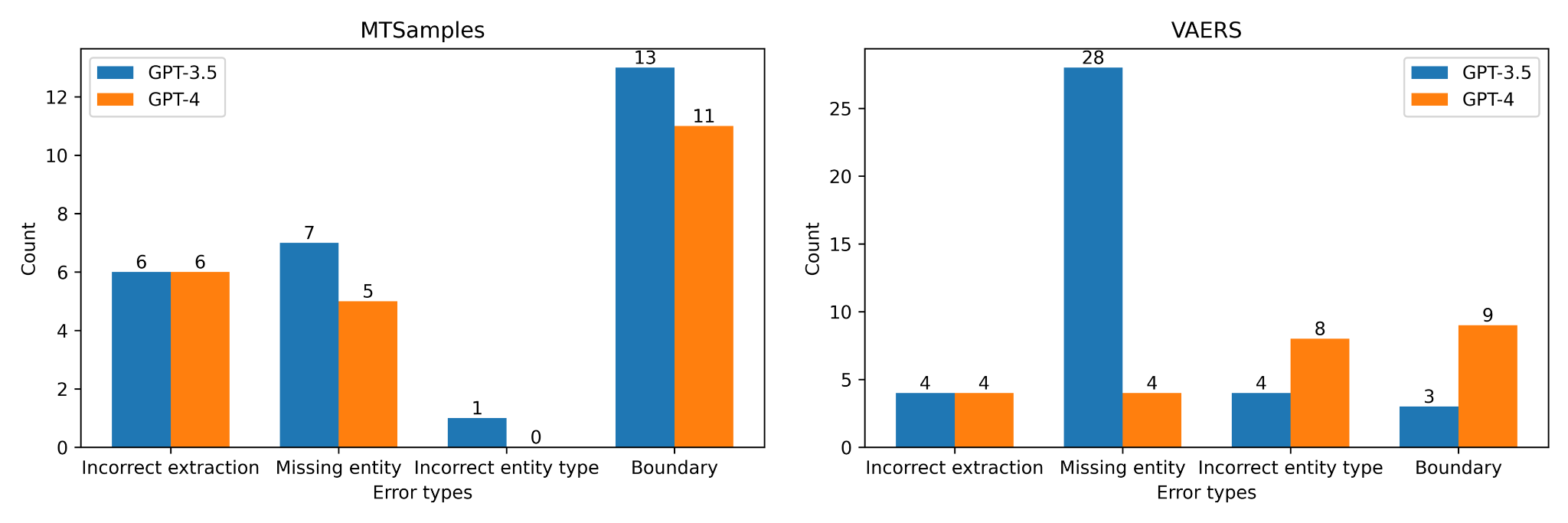}
 \caption{Performance comparison of GPT-3.5 and GPT-4 to BioClinicalBERT.}
 \label{fig:figure5_error_analysis}
\end{figure}

\subsection*{4 Discussion}
Our study hints at the as-yet unrealized potential of LLMs in clinical NER tasks by proposing a clinical task-specific prompt framework that incorporates annotation guidelines, error analysis-based instructions, and few-shot examples. We found that the performance of GPT models improved with the task-specific prompts. The best performance achieved by GPT-4 shows a competitive performance as that of BioClinicalBERT in the relaxed-match criteria. \\
LLMs are making paradigm-shifting changes in NLP research and development. Our finding shows a quick and easy path to build more generalizable clinical NER systems by leveraging LLMs. This will significantly change our current practice in clinical NLP. Traditionally, to build a machine learning or deep learning-based NER system for specific types of clinical entities, we have to build an annotated corpus of clinical documents, which is time-consuming and costly, as it often requires medical domain experts. Remarkably, our research shows that LLMs, devoid of further model training or fine-tuning, have exhibited exceptional performance. With merely 1- or 5-shot annotated samples, these models can achieve performance that is close to the fine-tuned models that require hundreds of training samples. This suggests a potential reduction in some of the costs associated with clinical NER system development, particularly in the areas of data annotation. However, it is important to note that this does not eliminate the need for expert input in creating annotation guidelines and in the initial phases of model training. While our study demonstrates that GPT models can achieve competitive performance with fewer annotated examples compared to traditional NLP systems, the role of subject matter experts remains crucial. Experts are needed to write precise annotation guidelines, perform initial annotations for error analysis and example generation, and validate the model's performance. Although the GPT models require fewer annotated instances, the costs associated with expert involvement, API usage, and running an LLM service should not be overlooked. A comprehensive comparison of resource requirements and costs between traditional NLP systems, word embedding models, and LLM-based systems would be valuable for future studies. This will provide a clearer understanding of the practical implications and feasibility of deploying LLMs in clinical NER tasks.\\
Moreover, our approach is generalizable – it shows consistent performance improvements across two different clinical NER tasks. The emergent abilities of LLMs \cite{wei2022emergent} have been further demonstrated in multiple clinical NER tasks here, indicating the feasibility of building one large model for diverse information extraction tasks in the medical domain, which is very appealing.  \\
With those changes in mind, an urgent need will be to re-design the workflow for developing clinical NER systems using LLMs. The prompt framework for those two clinical NER tasks is the first step toward this direction and it sheds some lights for several aspects that are worth considering. The first aspect is how to clearly define an information extraction task. Our experiments show specific annotation guidelines are very helpful, which indicates medical knowledge (either in a knowledge base or from human experts) are still critical in LLMs-based NER systems and how to obtain and represent task-specific knowledge in prompts need further investigation. We also demonstrated that supplying annotated examples is effective for performance improvement. Nevertheless, how to select informative and representative samples have not been investigated in this study and other advanced few-shot learning algorithms could be explored. \\
Another important issue is evaluation. In this study, we instructed GPT models to output entities following traditional NER approaches so that we can evaluate them using the previous evaluation scripts. However, we would argue that the current evaluation schema for NER may not be ideal for LLMs-based systems. GPT models, due to their generative nature and extensive pre-training on diverse text corpora, exhibit a nuanced understanding of context and language structure. This enables them to interpret and generate text in a way that sometimes extends beyond the strict boundaries of predefined entity classes. For instance, GPT models often recognized lab tests with abnormal values (e.g., “a blood sugar level of 40” or “white blood cell count of 23,500”) as medical problems. While this interpretation is contextually relevant and clinically meaningful, it deviates from the strict entity definitions used in our evaluation, leading to apparent mismatches. Therefore, a better evaluation schema would be needed to assess LLMs performance more accurately. \\
Despite the promising results, our study has some limitations. First, we limited LLMs to GPT models in this study. In future, we will include other popular LLMs such as LLaMA and Falcon \cite{touvron2023llama,touvron2023llama2,almazrouei2023falcon}. Second, our few-shot learning approaches were relatively simple, and we plan to investigate other approaches such as the chain-of-thoughts method \cite{chen2022program,sun2023enhancing,fu2022complexity}, hoping to yield better results. \\

\subsection*{5 Conclusion}
This is one of the first studies that systematically investigated GPT models for clinical NER via prompt engineering. In this study, we proposed a clinical task-specific prompt framework by incorporating annotation guidelines, error analysis-based instructions, and annotated samples via few-shot learning, and our evaluation on two clinical NER tasks show that the GPT-4 model with our proposed prompts achieved close performance as the state-of-the-art BioClinicalBERT model. The best performance achieved by GPT-4 with 5-shot learning did not work as well as the BioClinicalBERT model on MTSamples and VAERS datasets. Nevertheless, considering almost no training data was used in GPT, its performance is impressive hints the potential of LLMs in clinical NER tasks. While the results demonstrate a promising direction, they also underscore the need for further refinement and development before LLMs can consistently outperform established models like BioClinicalBERT in these specific applications. 

\subsection*{Funding Statement}
This work was supported by NIH grant number R21EB029575, R21AI164100, R01LM011934, 1K99LM01402,
R01AG066749, R01AG066749-03S1, R01LM013712, and U01TR002062; NIA grant number 1RF1AG072799, \\
1R01AG080429; CPRIT grant number RR180012; NSF grant number 2124789.

\subsection*{Conflict of Interest}
Dr. Hua Xu and Dr. Jingcheng Du have research-related financial interests at Melax Technologies Inc. 

\subsection*{Data Availability}
Our code and datasets are available at Github \footnote{\url{https://github.com/BIDS-Xu-Lab/Clinical_Entity_Recognition_Using_GPT_models}}.

\makeatletter
\renewcommand{\@biblabel}[1]{\hfill #1.}
\makeatother

\bibliographystyle{vancouver}
\bibliography{amia}  

\newpage

\subsection*{Supplementary Information:}
\subsection*{1 Supplementary Materials}
\subsection*{1.1 Complete prompts for two datasets}
\subsection*{1.1.1 The 2010 i2b2 concept extraction task}
\#\#\# Task\\
Your task is to generate an HTML version of an input text, marking up specific entities related to healthcare. The entities to be identified are: 'medical problems', 'treatments', and 'tests'. Use HTML \textless span \textgreater  tags to highlight these entities. Each \textless span \textgreater  should have a class attribute indicating the type of the entity.\\
\\
\#\#\# Entity Markup Guide\\
Use \textless span class="problem" \textgreater  to denote a medical problem.\\
Use \textless span class="treatment" \textgreater  to denote a treatment.\\
Use \textless span class="test" \textgreater  to denote a test.\\
Leave the text as it is if no such entities are found.\\
\\
\#\#\# Entity Definitions\\
Medical Problems are defined as: phrases that contain observations made by patients or clinicians about the patient’s body or mind that are thought to be abnormal or caused by a disease. They are loosely based on the UMLS semantic types of pathologic functions, disease or syndrome, mental or behavioral dysfunction, cellormolecular dysfunction, congenital abnormality, acquired abnormality, injury or poisoning, anatomic abnormality, neoplastic process, virus/bacterium, sign or symptom, but are not limited by UMLS coverage.\\
Treatments are defined as: phrases that describe procedures, interventions, and substances given to a patient in an effort to resolve a medical problem. They are loosely based on the UMLS semantic types therapeutic or preventive procedure, medical device, steroid, pharmacologic substance, biomedical or dental material, antibiotic, clinical drug, and drug delivery device.  Other concepts that are treatments but that may not be found in UMLS are also included. Treatments that a patient had, will have, may have in the future, or are explicitly mentioned that the patient will not have are all marked as treatments.\\
Tests are defined as: phrases that describe procedures, panels, and measures that are done to a patient or a body fluid or sample in order to discover, rule out, or find more information about a medical problem. They are loosely based on the UMLS semantic types laboratory procedure, diagnostic procedure, but also include instances not covered by UMLS.\\
\\
\#\#\# Annotation Guidelines\\
Only complete noun phrases (NPs) and adjective phrases (APs) should be marked. Terms that fit concept semantic rules, but that are only used as modifiers in a noun phrase should not be marked.\\
Include all modifiers with concepts when they appear in the same phrase except for assertion modifiers.\\
You can include up to one prepositional phrase (PP) following a markable concept if the PP does not contain a markable concept and either indicates an organ/body part or can be rearranged to eliminate the PP (we later call this the PP test).\\
Include articles and possessives.\\
Conjunctions and other syntax that denote lists should be included if they occur within the modifiers or are connected by a common set of modifiers. If the portions of the list are otherwise independent, they should not be included.  Similarly, when concepts are mentioned in more than one way in the same noun phrase (such as the definition of an acronym or where a generic and a brand name of a drug are used together), the concepts should be marked together.\\
Concepts should be mentioned in relation to the patient or someone else in the note. Section headers that provide formatting, but that are not specific to a person are not marked.\\
\\
\#\#\# Error-analysis-based Guidelines:\\
Vital signs or vital signs with abnormal readings should be annotated as tests.\\
Medical specialists, services, or healthcare facilities should not be annotated, even if they might seem to fit into the categories of 'tests', 'treatments', or 'medical problems'. These entities are part of the healthcare delivery system and do not directly denote a test, treatment, or medical problem.\\
Consultation procedures should not be considered as tests.\\
\\
\#\#\# Examples \\
Example Input1: At the time of admission , he denied fever , diaphoresis , nausea , chest pain or other systemic symptoms .  \\
Example Output1: At the time of admission , he denied \textless span class="problem" \textgreater fever\textless /span \textgreater  , \textless span class="problem" \textgreater diaphoresis\textless /span \textgreater  , \textless span class="problem" \textgreater nausea\textless /span \textgreater  , \textless span class="problem" \textgreater chest pain\textless /span \textgreater  or other systemic symptoms .  \\
Example Input2: He had been diagnosed with osteoarthritis of the knees and had undergone arthroscopy years prior to admission .  \\
Example Output2: He had been diagnosed with \textless span class="problem" \textgreater osteoarthritis of the knees\textless /span \textgreater  and had undergone \textless span class="test" \textgreater arthroscopy\textless /span \textgreater  years prior to admission .  \\
Example Input3: After the patient was seen in the office on August 10 , she persisted with high fevers and was admitted on August 11 to Cottonwood Hospital .  \\
Example Output3: After the patient was seen in the office on August 10 , she persisted with \textless span class="problem" \textgreater high fevers\textless /span \textgreater  and was admitted on August 11 to Cottonwood Hospital .  \\
Example Input4: HISTORY OF PRESENT ILLNESS : The patient is an 85 - year - old male who was brought in by EMS with a complaint of a decreased level of consciousness .  \\
Example Output4: HISTORY OF PRESENT ILLNESS : The patient is an 85 - year - old male who was brought in by EMS with a complaint of \textless span class="problem" \textgreater a decreased level of consciousness\textless /span \textgreater  .  \\
Example Input5: Her lisinopril was increased to 40 mg daily .  \\
Example Output5: \textless span class="treatment" \textgreater Her lisinopril\textless /span \textgreater  was increased to 40 mg daily .  \\
\\
\#\#\# Input Text: {}\\
\#\#\# Output Text:\\

\subsection*{1.1.2 The nervous system disorder-related event extraction task}

\#\#\# Task\\
Your task is to generate an HTML version of an input text, marking up specific entities related to healthcare. The entities to be identified are: 'investigations', 'nervous adverse events', 'other adverse events', and 'procedures'. Use HTML \textless span \textgreater  tags to highlight these entities. Each \textless span \textgreater  should have a class attribute indicating the type of the entity.\\
\\
\#\#\# Entity Markup Guide\\
Use \textless span class="investigation" \textgreater  to denote an investigation.\\
Use \textless span class="nervous\_AE" \textgreater  to denote a nervous adverse event.\\
Use \textless span class="other\_AE" \textgreater  to denote an other adverse event.\\
Use \textless span class="procedure" \textgreater  to denote a procedure.\\
If no entity found, leave the text as it is.\\
\\
\#\#\# Entity Definitions\\
Investigation includes typical lab tests or examinations in the report, such as physical examination, oxygen saturation, electromyogram, etc.\\
Nervous adverse event includes typically nervous system-related problems, such as guillain-barré syndrome, ataxia, areflexia, hypoaesthesia, paraesthesia, dizziness, headache and other nervous system disorders.\\
Other adverse event includes medical problems that are assigned to other MedDRA SOCs, including gastrointestinal disorders, cardiac disorders, psychiatric disorders, musculoskeletal and connective tissue disorders, etc.\\
Procedure includes non-medical problem events such as individual immunization complications or related medical events (each immunization should be marked separately), surgeries such as catheter placement, hospitalization, emergence care, intubation, etc. A procedure refers to a specific medical or surgical activity carried out to diagnose, treat, or monitor a condition. Routine care activities or general healthcare administration such as 'sick call', 'doctor's visit', 'general checkup', etc. without a specific associated procedure or event should not be considered as a procedure. Note that 'vaccines administered' in absence of any complications or related medical events should not be considered a procedure.\\
Please note that in the case of negation where a certain adverse event, investigation, or procedure is clearly indicated NOT to have occurred (e.g., 'No bowel or bladder symptoms'), do not mark the entity.\\
\\
\#\#\# Annotation Guidelines\\
Only annotate events that already occurred (i.e., occurred before the diagnosis of GBS).\\
When annotating events related to Flu-GBS, do not include prepositions including modifiers of the event.\\
Separate events in discontinuous segments.\\
When annotating events, more generalized events should not be annotated.\\
When annotating events related to symptom improvement / progress or negation events, the following guideline should be used. In the case where the patient reported a specific adverse event first, and then reported improvement / progress of the adverse event, we should annotate it as an improved symptom. However, we do NOT need to annotate the negation of a symptom which the patient never reported before.\\
Events reported as history (events that did not happen to the reporting patient) should be annotated. Family history is important for risk prediction and may be included as a baseline information (e.g., for statistical analysis).\\
Some VAERS reports have duplicate events reported. For example, the same events / text are repeated twice in the report. The case we are interested in, is the recurrence of some adverse event, i.e., it requires the adverse event appears, then disappear, and then come back. In this case it should definitely be annotated twice. Additionally, we need to annotate the relief/improvement of the event if it is mentioned in the report. When no such information to decide whether it is a recurrence, the principle is that if there are multiple time stamps of the same event, we annotate it twice, if not, we can just keep one record.\\
\\
\#\#\# Error-analysis-based Guidelines:\\
When annotating events related to hospital admissions, transfers, or discharges, consider them as procedures. Specifically, annotate the words 'hospital', 'rehabilitation center', or any other healthcare facility involved in the patient's care as a procedure.\\
All abnormal symptoms should be considered as adverse events.\\
\\
\#\#\# Examples \\
Example Input1: Received flu shot 11 / 1 / 06 .  \\
Example Output1: Received \textless span class="procedure" \textgreater flu shot\textless /span \textgreater  11 / 1 / 06 .  \\
Example Input2: 1 / 28 / 05 PM : ascending redness left elbow then from fingertips .  \\
Example Output2: 1 / 28 / 05 PM : \textless span class="other\_AE" \textgreater ascending redness left elbow then from fingertips\textless /span \textgreater  .  \\
Example Input3: Unable to stand due to severe ataxia .  \\
Example Output3: \textless span class="nervous\_AE" \textgreater Unable to stand\textless /span \textgreater  due to severe \textless span class="nervous\_AE" \textgreater ataxia\textless /span \textgreater  .  \\
Example Input4: At 4 AM on 12 - 16 - 11 got up again to go to the bathroom and on the way out my right leg gave out from under me again and my husband saw me and tried to help me and then both legs wouldn ' t work .  \\
Example Output4: At 4 AM on 12 - 16 - 11 got up again to go to the bathroom and on the way out my \textless span class="nervous\_AE" \textgreater right leg gave out\textless /span \textgreater  from under me again and my husband saw me and tried to help me and then \textless span class="nervous\_AE" \textgreater both legs wouldn ' t work\textless /span \textgreater  .  \\
Example Input5: Seen by neurologist and diagnosed with Guillain Barre Syndrome .  \\
Example Output5: Seen by neurologist and diagnosed with \textless span class="nervous\_AE" \textgreater Guillain Barre Syndrome\textless /span \textgreater  .  \\
\\
\#\#\# Input Text: {}\\
\#\#\# Output Text:\\

\subsection*{1.2 Learning Curve of BioClinicalBERT on the validation sets}
To provide additional insights into model training and validation, we conducted a learning curve analysis for both the MTSamples and VAERS datasets using BioClinicalBERT. The learning curves, depicted in Figures S1, illustrate the model's performance over epochs on the validation set. For MTSamples, the F1 score improved sharply in the initial epochs, and plateaued around epoch 5. In the case of VAERS, the improvement in F1 was also sharp in the beginning and leveled off near epoch 4, maintaining a consistent score thereafter. These trends suggest that the model reached its performance capacity quickly and did not exhibit signs of overfitting, as evidenced by the stable F1 scores beyond the plateau point. 

\renewcommand{\figurename}{Supplementary Figure S}
\setcounter{figure}{0}
\begin{figure}[H]
  \centering
    \includegraphics[width=0.8\textwidth]{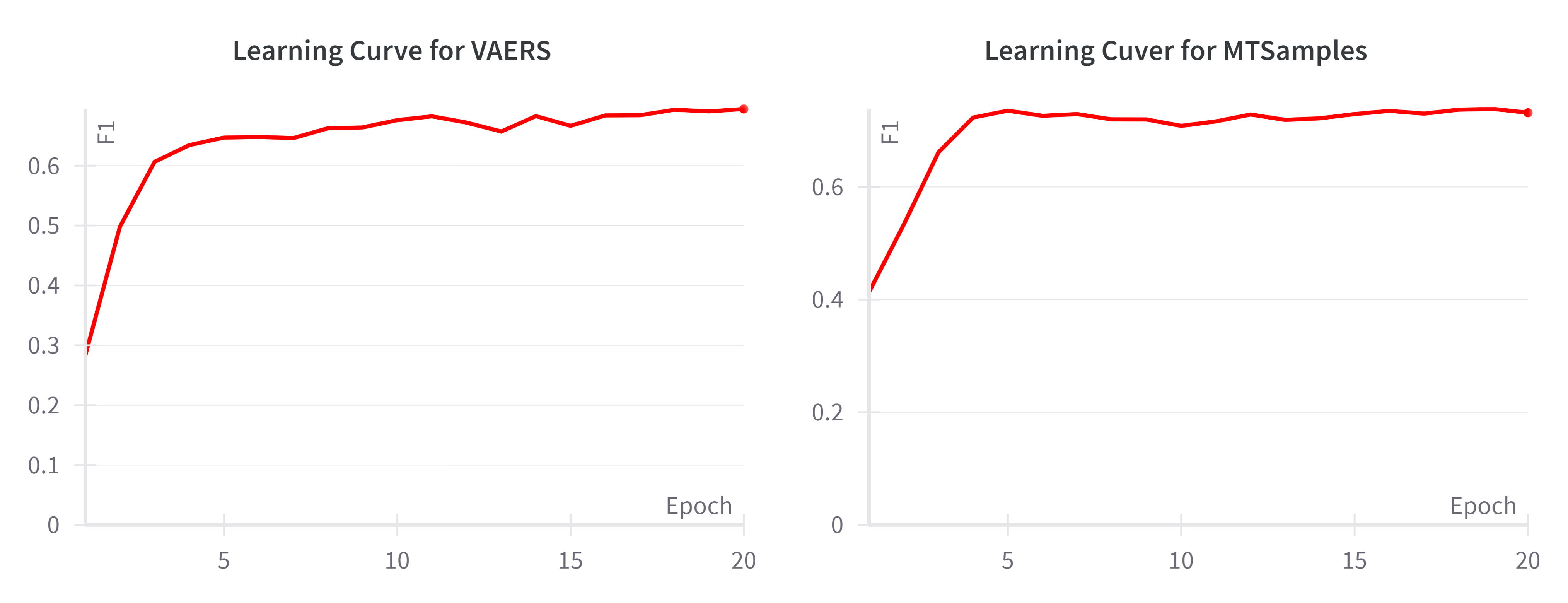}
 \caption{Learning curve of BioClinicalBERT on validation sets across epochs for MTSamples and VAERS datasets}
 \label{fig:figure6_learning_curve}
\end{figure}

\end{document}